\documentclass[pdflatex,sn-mathphys-num]{sn-jnl}

\usepackage{graphicx}%
\usepackage{multirow}%
\usepackage{amsmath,amssymb,amsfonts}%
\usepackage{amsthm}%
\usepackage{mathrsfs}%
\usepackage[title]{appendix}%
\usepackage{xcolor}%
\usepackage{textcomp}%
\usepackage{manyfoot}%
\usepackage{booktabs}%
\usepackage{algorithm}%
\usepackage{algorithmicx}%
\usepackage{algpseudocode}%
\usepackage{listings}%

\theoremstyle{thmstyleone}%
\theoremstyle{thmstyletwo}%

\theoremstyle{thmstylethree}%

\begin{document}

\title[Optimism, Expectation or Sarcasm?]{Optimism, Expectation, or Sarcasm? Multi-Class Hope Speech Detection in Spanish and English}


\author*[1]{\fnm{Sabur} \sur{Butt}}\email{saburb@tec.mx}

\author[2]{\fnm{Fazlourrahman} \sur{Balouchzahi}}\email{fbalouchzahi2021@cic.ipn.mx}

\author[3]{\fnm{Ahmad} \sur{Imam Amjad}}\email{ahmadimamamjad@gmail.com}

\author[4]{\fnm{Maaz} \sur{Amjad}}\email{maaz.amjad@ttu.edu}

\author[1]{\fnm{Hector} \sur{G. Ceballos}}\email{ceballos@tec.mx}

\author[5]{\fnm{Salud María} \sur{Jiménez-Zafra}}\email{sjzafra@ujaen.es}

\affil*[1]{\orgdiv{Institute for the Future Education}, \orgname{Tecnológico de Monterrey }, \orgaddress{ \city{Monterrey}, \country{Mexico}}}

\affil[2]{\orgdiv{Centro de Investigación en Computación}, \orgname{IPN}, \orgaddress{\city{Ciudad de México}, \country{Mexico}}}

\affil[3]{\orgdiv{Department of Computer Science}, \orgname{The University of Punjab},\orgaddress{ \city{Lahore}, \country{Pakistan}}}

\affil[4]{\orgdiv{Department of Computer Science}, \orgname{Texas Tech University}, \orgaddress{\city{Lubbock}, \state{Texas}, \country{United States}}}

\affil[5]{\orgdiv{Computer Science Department, SINAI, CEATIC}, \orgname{Universidad de Jaén}, \orgaddress{, \country{Spain}}}


\abstract{Hope is a complex and underexplored emotional state that plays a significant role in education, mental health, and social interaction. Unlike basic emotions, hope manifests in nuanced forms ranging from grounded optimism to exaggerated wishfulness or sarcasm, making it difficult for Natural Language Processing systems to detect accurately. This study introduces PolyHope V2, a multilingual, fine-grained hope-speech dataset comprising over 30,000 annotated tweets in English and Spanish. This resource distinguishes between four hope subtypes—Generalized, Realistic, Unrealistic, and Sarcastic—and enhances existing datasets by explicitly labeling sarcastic instances. We benchmark multiple pre-trained transformer models and compare them with large language models (LLMs) such as GPT-4 and Llama 3 under zero-shot and few-shot regimes. Our findings show that fine-tuned transformers outperform prompt-based LLMs, especially in distinguishing nuanced hope categories and sarcasm. Through qualitative analysis and confusion matrices, we highlight systematic challenges in separating closely related hope subtypes. The dataset and results provide a robust foundation for future emotion recognition tasks that demand greater semantic and contextual sensitivity across languages.}




\maketitle

\section{Introduction} 
Recent improvements in Natural Language Processing (NLP) have enhanced applications in sentiment analysis, mental health assessments, social media monitoring, and educational platforms~\cite{poria2016deeper, cambria2013new, bustos2021emotion, gomez2023emotion, mishra2025unveiling}. Despite recent progress, a persistent challenge in emotion recognition lies in identifying subtle and complex emotions, particularly hope, which is often overlooked in standard emotion taxonomies~\cite{mohammad2021sentiment}. Unlike overt emotions such as anger, fear, sadness, or joy, hope is an inherently ambiguous emotion that exhibits differently across various contexts, and carries multiple emotional dimensions \cite{buechel2022emobank, chatterjee2019semeval}. Therefore, traditional NLP techniques struggle to capture subtle emotions and highlight the need for more refined, context-aware approaches to accurately detect nuanced emotional expressions, such as hope~\cite{balouchzahi2023polyhope}.

Accurate detection of hope speech unlocks substantial benefits across real-world NLP applications, such as enhancing student engagement and success \cite{xie2019trends}, understanding patients' mental states \cite{resnik2015beyond, balouchzahi4930517analysis}, and Non-medical prescription drug use (NPMDU). For example, within online education, instructors can leverage these systems as personalized support tools to accurately identify expressions of hope to better understand students' motivations, interests, and emotional barriers—ultimately enabling targeted interventions~\cite{xie2019trends}. Similarly, mental health professionals can leverage the natural progression of hope speech to better understand patients' mental states, therapeutic outcomes, and concealed psychological factors over time. Research emphasizes that accurately interpreting psychological states requires identifying positive emotional expressions, particularly hope and optimism \cite{resnik2015beyond}. As a result, NLP systems significantly benefit from incorporating hope detection technologies, since integrating positive emotional indicators alongside negative sentiments enables more comprehensive, nuanced, and effective analytic models.

Traditional NLP techniques struggle to detect hope due to its complex, dual nature as defined by Lazarus \cite{lazarus1991emotion}. According to Lazarus, hope consists of two essential components: (i) a specific desire for positive outcomes, and (ii) an overall optimistic view of future events. The inherent contradictions within hope result in complicated textual interpretations, causing individual textual markers to be variably perceived as genuine goals, imaginary objectives, or sardonic remarks. Consequently, expressions intended to reflect hopeful thinking may inadvertently suggest agitation, depending on their textual context. For example, the sentence ``I hope things finally change around here" reveals optimism but also hints at frustration, depending on how it is used in the text. Most emotion recognition systems trained on simplified affective categories (e.g., Ekman’s six basic emotions) tend to fail when processing hope, as a generic positive sentiment or as non-committal discourse, failing to grasp its subtleties \cite{ekman1992argument, buechel2022emobank}. Thus, emotion recognition systems trained on basic emotion sets fail to correctly identify hope and often misclassify it either as a simple positive emotion or as a neutral, non-committal state.

Additionally, recognizing hope is further complicated by sarcasm, as sarcastic expressions frequently employ positive language to communicate negative sentiments. The mismatch between literal meanings and intended emotions makes it challenging for machine learning and deep learning classifiers to distinguish genuinely hopeful statements from their sarcastic counterparts. For instance, sarcasm occurs when someone says, ``Great, I really hope this mess gets worse," even though the words themselves convey hope, but the semantic context reveals sarcastic negativity \cite{riloff2013sarcasm}.

The semantic incongruity between lexical content and emotional meaning makes emotion detection more challenging. The combination of figurative language and sarcasm complicates consistent emotion classification for computers since they primarily work with superficial word analysis instead of contextual interpretation \cite{poria2016deeper}. Accurate context-aware emotion detection primarily depends on extensive, diverse datasets and sophisticated models \cite{cambria2017sentiment}. However, there is a notable lack of annotated datasets explicitly related to hope, and existing datasets often either entirely omit hope or classify it merely as a subset of general positive emotions that cause incorrect predictions. Therefore, more comprehensive annotated datasets are necessary that explicitly capture hope through genuinely positive sentiments, unfounded expectations, and sarcastic expressions.

In this study, we bridge the gap between emotion recognition and pragmatic language understanding and present a bilingual annotated hope speech dataset designed for hope detection in both English and Spanish. We explicitly annotate sarcasm within hope expressions and create rich annotation guidelines for different hope categories. We extend~\cite{balouchzahi2023polyhope} hope categories (i) generalized hope, (ii) realistic hope, (iii) unrealistic hope or wishful thinking, and (iv) sarcasm. We develop sarcasm-resilient emotion classification models as a baseline to detect and evaluate the performance of the hope detection system in English and Spanish. This dataset can be used for advancing cross-linguistic emotion recognition research. 

\section{Literature Review} 
NLP has advanced beyond basic sentiment classification (positive, negative, or neutral) toward identifying more nuanced emotional states, including the subtle and underexplored emotion of hope \cite{cambria2017sentiment, mohammad2021sentiment}. While emotions such as joy, fear, and regret have received substantial attention, hope remains comparatively underresearched, despite its significant role in mental health support
\cite{resnik2015beyond}, education \cite{xie2019trends}, online communication, and social programs \cite{ullah2024hope}. Hope consists of two distinct components that contribute to its complexity: (i) a general optimism inherent in individuals, and (ii) a specific expectation for positive outcomes \cite{lazarus1991emotion}. Given its unique semantic, emotional, and contextual characteristics, recent research in hope speech detection has increasingly treated hope as an independent emotional category rather than merely grouping it under general positive emotions.

\subsection{Methodologies for Hope Speech Detection}~\label{Section: LR Method}

Different methodologies have been proposed for detecting hope speech and addressing challenges related to context, semantics, and multilingual content. Various studies have explored techniques to identify hope expressions within social media and textual documents \cite{strapparava2008learning, palakodety2020hope}. For instance, Balouchzahi et al. \cite{balouchzahi2023polyhope} introduced ``PolyHope," an NLP system that operates at semantic and contextual levels to detect nuanced expressions of hope. Similarly, another study \cite{sidorov2023regret} leveraged transformer-based models to differentiate between expressions of hope and regret, and highlighted complexities in jointly modeling these emotionally similar pairs. Despite these advancements, current methodologies lack in distinguishing hope expressions from sarcastic or superficially hopeful messages, particularly on social media platforms (e.g., X (previously Twitter), and Reddit). Deep learning models provide poor results because they often misinterpret pragmatic signals and contextual discourse markers \cite{taboada2016sentiment}. These limitations underline the need to incorporate pragmatic and contextual awareness to improve model accuracy. Moreover, addressing these contextual challenges becomes even more critical when extending hope detection methodologies across multilingual and cross-cultural datasets.

A growing body of work has explored multilingual and cross-lingual hope speech detection, recognizing its importance in linguistically diverse societies. Chakravarthi et al. \cite{chakravarthi2020hopeedi, chakravarthi2022overview} introduced multilingual approaches to detect hope speech in English, Tamil, Malayalam, and Kannada. The authors explicitly emphasized equality and diversity and inclusion (EDI) to detect hope speech. Nevertheless, their models did not sufficiently account for distortions caused by sarcasm or unrealistic expectations embedded within hopeful expressions. Similarly, researchers such as Arunima et al. \cite{arunima2021ssn_dibertsity} have contributed to shared tasks like Hope Speech Detection for Equality, Diversity, and Inclusion (HopeEDI), which expanded datasets and evaluation benchmarks in multiple Indian languages. Additional efforts have targeted Urdu datasets, with studies exploring the transferability of hope speech models across languages and the need for language-specific semantic resources \cite{balouchzahi2025urduhope, ponnusamy2024vel}. Chakravarti \cite{chakravarthi2022multilingual} addressed hope speech detection in Tamil, English, and Malayalam, highlighting classification challenges due to code-mixed data. Moreover, Malik et al. \cite{malik2023multilingual} explored a joint multilingual and translation-based approach for hope speech detection in English and Russian using a fine-tuned Russian-RoBERTa model, achieving 94\% accuracy and 80.24\% F1-score, though multiclass classification was not addressed.
However, most of these models do not explicitly address the confounding effects of sarcasm, exaggeration, or unrealistic optimism expressed through hopeful language—factors that often distort detection accuracy in real-world applications. Moreover, research on low-resource and code-mixed language scenarios remains limited, although recent work has begun to tackle these challenges through cross-lingual transfer learning and zero-shot classification approaches \cite{thangaraj2024cross}.

Moreover, the integration of sarcasm detection and contextual awareness into current lexical and dictionary-based hope detection methods often underperformed in analyzing unstructured, informal textual data. Therefore, integrating pragmatic, contextual, and multilingual perspectives to ensure accurate and reliable emotion recognition outcomes is crucial.

\subsection{Existing Datasets for Hope Speech Detection} 

Several datasets have been introduced to facilitate advancements in hope speech detection in various languages, such as
Spanish \cite{garcia2023hope, garcia2024overview}, Bengali \cite{nath2023bonghope}. For example, the ``HopeEDI" dataset \cite{chakravarthi2020hopeedi} introduced a multilingual corpus specifically targeting hopeful expressions related to equality, diversity, and inclusion (EDI). Additionally, the ``IberLEF" Hope datasets \cite{garcia2023hope, garcia2024overview} extended hope speech detection into Spanish, and emphasized annotations around social causes and community identities. Another dataset, ``BongHope" \cite{nath2023bonghope}, presented a Bengali corpus for hope speech detection. Furthermore, the HOPE shared tasks at IberLEF provided comprehensive benchmarks and detailed insights into effective methodologies for annotating hope datasets \cite{garcia2024overview, jimenez2023overview}. In parallel, the HASOC-Hope subtrack, part of the FIRE shared tasks, introduced additional multilingual datasets covering Hindi, Malayalam, and Tamil, further enriching the research landscape in this area \cite{mandl2020overview}). Finally, the LREC Hope Speech dataset offers a multilingual resource tailored for cross-lingual and zero-shot hope speech detection, with a special focus on low-resource language scenarios \cite{pelicon2021zero}.

However, most existing datasets either categorize hope speech through a simplistic binary labeling scheme (hope vs. non-hope) or do not annotate nuanced subcategories such as generalized optimism, unrealistic hope, or sarcastic hope. Additionally, few datasets explicitly annotate sarcasm, which limits their usefulness in training or evaluating models capable of differentiating genuine hope from sarcastic expressions. Moreover, annotation schemas typically lack cross-linguistic alignment, which poses challenges for developing robust multilingual models. As a result, there remains a notable gap in available datasets that provide both fine-grained hope annotations and explicit sarcasm labeling.

\subsection{Sarcasm in Hope Speech: A Complex Challenge}

The detection of hope speech is highly impeded by the presence of sarcasm. Positive sentimental language through rhetorical expression functions as sarcasm to convey negative meanings while possessing a humorous or mocking quality. For example, ``This one needs to go well as it happened previously". This hopeful-sounding statement contains semantic skepticism through ironic speech. The model detection of hopeful speech can fail due to sarcastic statements that resemble hopeful statements. The improper evaluation of sentiment features occurs when basic analysis methods, such as positive polarity measurements, operate independently from detailed contextual analysis. 

The task of sarcasm detection is challenging because it requires contextual information, tonal cues, and background understanding \cite{cambria2017sentiment}. Research on sarcasm detection within NLP has approached this problem in four ways: (i) Contextual embeddings (e.g., BERT, RoBERTa)~\cite{srivastava2020novel}, (ii) Multitask learning frameworks (to jointly learn emotion and sarcasm)~\cite{chauhan2020sentiment}, (iii) Contrastive learning approaches~\cite{jia2024debiasing}, and (iv) Discourse-aware models~\cite{dong2020transformer} that incorporate surrounding sentences or conversational history. Yet sarcasm-aware hope detection remains an unexplored intersection, and no existing hope detection dataset provides labeled sarcastic examples, as emphasized by Balouchzahi et al. \cite{balouchzahi2023polyhope} in their analysis of NPMDU-related textual data. To address this challenge, we annotate hope expressions along with sarcasm indicators and provide a richer, more realistic representation of hope in social discourse. Such annotation is critical for building models that can distinguish between sincere emotional optimism and sarcastic cynicism, as such a distinction is particularly important in domains like education, counseling, and mental health.

\section{Dataset}

In this work, we present PolyHope V2, a multilingual and fine-grained hope speech dataset consisting of around 30,000 annotated tweets in both English and Spanish. PolyHope~V2 brings together the original English‐language PolyHope tweets~\cite{balouchzahi2023polyhope}, a newly harvested Spanish counterpart, and an additional sarcasm layer to create a multilingual, multi‑label resource for modelling hope on social media. Altogether the corpus contains 9,515 English and 20,442 Spanish tweets, each message annotated at the tweet level for hope, sarcasm, and the related fine‑grained categories described below. The next subsections outline how the English, Spanish, and sarcasm components were collected, cleaned, and aligned. The complete dataset is publicly available on the web page of a shared task that we have organized to evaluate it and make it available to the scientific community\footnote{\url{https://www.codabench.org/competitions/5509/}}. 

\subsection{English}
We relied on the PolyHope corpus~\cite{balouchzahi2023polyhope} for both binary and multi‑class modeling. \citet {balouchzahi2023polyhope} used Twitter to collect hope‑related instances, gathering roughly 100,000 English tweets posted between January and June 2022, a period dominated by discussions of women’s reproductive rights, Black civil‑rights issues, religion, and mainstream politics. After discarding incomplete or duplicate texts, the pool dropped to just under 70,000 tweets. Filtering out messages with fewer than ten tokens reduced the set to fewer than 50,000, and the subsequent removal of retweets left about 23,000 unique, original tweets. These were shuffled to avoid topical clustering, and a random sample of 10,000 tweets was selected for manual annotation. A small number were eliminated during quality control, so the final annotated subset is slightly smaller than the initial sample, but still provides a substantial and balanced foundation for modelling hope phenomena in English social‑media discourse. The further details on annotation and pre-processing are provided in the dataset paper~\cite{balouchzahi2023polyhope}.

\subsection{Spanish}
For Spanish, we rely on the corpus introduced by~\citet{sidorov2024mind} for both binary and multi‑class hope detection. The authors translated the PolyHope English trigger words into Spanish (e.g., hope → esperanza, wish → ojalá, believe → creo, dream → sueño) and asked native speakers to check that the list covered all common variants. Using these cues they first collected 82,725 keyword‑matching tweets, then added the 50,000 most recent Spanish tweets available in March 2022, yielding 132,725 raw items. After standard preprocessing—de‑identifying user mentions and URLs, removing duplicates and retweets, and discarding messages with fewer than five tokens or only emojis—the dataset was reduced to roughly 33,300 unique, content‑rich tweets. After pre-processing, only 35,000 tweets were left, and the remaining were excluded during the quality control and annotation phase. The further details on annotation and pre-processing are provided in the dataset paper~\cite{sidorov2024mind}. 

\subsection{Sarcasm for English and Spanish}
We expanded the English part of the dataset using two benchmark datasets in sarcasm detection iSarcasmEval~\cite{farha2022semeval} and Sarcasm Corpus V2~\cite{oraby2016creating}. The dataset is aligned with our task as they are also collected using Twitter (Now X). We combined these datasets and used our Hope classifier (the best transformer model described in~\cite{balouchzahi2023polyhope} fine-tuned on the PolyHope dataset), filtered all instances of sarcasm that can be detected as Hope in any category. Similarly, we developed a deep learning classifier (the best transformer model described in~\cite{balouchzahi2023polyhope} fine-tuned on the Sarcasm dataset) and filtered all instances of potential sarcasm in other classes of Hope, i.e., unrealistic hope. 

The Spanish instances of Sarcasm were synthetically generated using GPT-4o, by translating the English instances and keeping the sarcastic tone in Spanish. The new labels were then verified using two expert annotators. In the binary setting we merged every tweet annotated as ``sarcasm” with the broader “not‑hope” category, producing a clean dichotomy between hopeful and non‑hopeful discourse. This decision reflects the pragmatic observation that sarcastic posts rarely convey genuine hope; rather, they typically critique or ridicule the prospect of a positive outcome. The grouping of sarcasm with no hope, therefore, sharpens the semantic boundary the classifier must learn, reduces class imbalance, and minimizes the risk that ironic language will be misinterpreted as an expression of hope. 

For the multi‑class task, however, we preserved sarcasm as its own label alongside hope, not‑hope, and the other original classes provided by PolyHope. Retaining sarcasm allows the model to capture the distinctive lexical and pragmatic cues of ironic speech, which differ markedly from both sincerely hopeful and straightforwardly pessimistic language. Keeping it separate also facilitates finer‑grained downstream analyses—researchers can quantify how sarcastic framing interacts with sensitive topics such as reproductive rights or racial justice—while still leveraging the same annotated tweet set described in~\cite{balouchzahi2023polyhope} and~\cite{sidorov2024mind}. The examples of the data, in English and Spanish, can be seen in Table~\ref{tab:samples} and~\ref{tab:samples_es} respectively, while Table~\ref{tab:statistics} shows the statistics of the dataset.

\begin{table*}[!htbp]
    \centering
    \caption{\label{tab:samples}Sample English texts from new proposed version of PolyHope~V2 with Sarcasm.}
    \resizebox{\textwidth}{!}{%
    \begin{tabular}{l|c|c}
    \hline
    \multicolumn{3}{c}{\bf PolyHope} \\
    \hline
    \bf Text & \multicolumn{2}{c}{\bf Label}   \\ \hline
    \#USER\# You just can’t get out of your own way, can you? &\multicolumn{2}{c}{Not Hope}\\ \hline
    \begin{tabular}[c]{@{}l@{}}Congratulations \#USER\# bro for completing, my allah god bless u ahead\\a good and bright future inshallah... \#URL\#\end{tabular} & \multicolumn{2}{c}{Generalized Hope}\\\hline

    \begin{tabular}[c]{@{}l@{}}\#USER\# damn man i'm here just tryna be a nice friend and dis is how u treat me\\ i hope u choke with cold spaghetti\end{tabular} & \multicolumn{2}{c}{Unrealistic Hope}\\\hline

    \begin{tabular}[c]{@{}l@{}}It was really hard exam but fortunately I studied a little last night and \\I expect to pass the test\end{tabular} & \multicolumn{2}{c}{Realistic Hope}\\\hline
    
       \begin{tabular}[c]{@{}l@{}}Oh, I’m sure this meeting will solve all our problems!\end{tabular} & \multicolumn{2}{c}{Sarcasm}\\\hline

    \end{tabular}
    }
\end{table*}

\begin{table*}[!htbp]
    \centering
    \caption{\label{tab:samples_es}Sample Spanish tweets from PolyHope V2.}
    \resizebox{\textwidth}{!}{%
    \begin{tabular}{l|c|c}
    \hline
    \multicolumn{3}{c}{\bf PolyHope-Spanish} \\ \hline
    \bf Text & \multicolumn{2}{c}{\bf Label} \\ \hline
    \#USER\# Que gente de m\ldots{}sino saben no hablen. \\Tengo mel\ldots{} haya dado la teta a la melliza equivocada, entr el sueño y cans & \multicolumn{2}{c}{Not Hope} \\ \hline
    \#USER\# Uy! Ojalá.  Pero tiene que ser más como el \\\emph{Fight for NY} porque el \emph{Icon} era asquerosamente horrible. & \multicolumn{2}{c}{Generalized Hope} \\ \hline
    \#USER\# Aquí en la Mora y Piedra Azul seguimos esperando\\ la segunda fase de alumbrado público de sus calles \#USER\# \#USER\# & \multicolumn{2}{c}{Realistic Hope} \\ \hline
    Todo el maldito día trabajando en una presentación.\\ Ansío el día en que deje de ser una \emph{corporate slave}. & \multicolumn{2}{c}{Unrealistic Hope} \\ \hline
    Si el sonido de la lluvia afuera no fuera ya suficientemente\\ malo\ldots{} sería el fondo perfecto para mí mientras intento\\ estudiar. ¡Maravilloso! & \multicolumn{2}{c}{Sarcasm} \\ \hline
    \end{tabular}}
\end{table*}
\begin{table}[!htbp]
\caption{\label{tab:statistics}Statistics of the PolyHope~V2 dataset.}
\centering
\begin{tabular}{l|l|rr}
\hline
{ Task}                         & { Labels}            &English& Spanish  \\ \hline
                                & { Hope}              & 4,434 & 9,654     \\
\multirow{-2}{*}{{ Binary}}     & { Not Hope}          & 5,081 & 10,788      \\\hline
                                & { Not Hope}          &4,081& 9,788   \\
                                & { Sarcasm}          & 1,259 & 1259    \\
                                & { Generalized Hope}  &2,335 &   5,007    \\
                                & { Realistic Hope}    &982 & 2024       \\
\multirow{-5}{*}{{ Multiclass}} & { Unrealistic Hope}  &858 & 2364      \\\hline
\multicolumn{2}{c}{{ Total}}                           & 9,515 & 20,442 \\\hline
\end{tabular}

\end{table}

\section{Methodology}

This section details the baselines employed for hope speech and sarcasm detection within the PolyHope v.2 dataset. We explore two primary approaches: fine-tuning pre-trained Transformer models and leveraging Zero-Shot Learning (ZSL) and Few-Shot Learning (FSL) capabilities of Large Language Models (LLMs).

\subsection{Fine-tuning Pre-trained Transformer Models}

To establish strong and reproducible baselines for the proposed task, we adopted a fine-tuning approach using pre-trained Transformer models from the Hugging Face Transformers library~\footnote{\url{https://huggingface.co/models}}. This approach leverages transfer learning, enabling us to adapt models pre-trained on large corpora to our task-specific dataset. We fine-tuned several Transformer architectures, including Roberta, Albert, Electra, and DistilBERT, to evaluate their performance on both binary (Hope vs. Not Hope) and multiclass (Not Hope, Generalized Hope, Realistic Hope, Unrealistic Hope, and Sarcasm) classification tasks. 

\subsubsection{Experimental Setup}

We implemented a stratified 5-fold cross-validation strategy using the \texttt{simpletransformers} library~\footnote{\url{http://simpletransformers.ai/}} to ensure robust and generalizable evaluation. This is crucial to mitigate potential biases and provide reliable performance estimates. Within each fold, we executed the following steps:

\begin{enumerate}
    \item \textbf{Model Initialization:} We instantiated a \texttt{ClassificationModel} from the \texttt{simpletransformers} library, specifying the pre-trained architecture and model variant.
    \item \textbf{Fine-tuning Procedure:} The instantiated model was then fine-tuned on the training partition. We employed early stopping based on the development set performance to prevent overfitting.
    \item \textbf{Evaluation Metrics:} After fine-tuning, the model's performance was assessed on the held-out test split. We report accuracy, weighted precision, weighted recall, weighted F1-score, macro precision, macro recall, and macro F1-score. We emphasize weighted and macro averages due to the imbalanced nature of the dataset, providing a comprehensive performance profile across all classes.
    \item \textbf{Qualitative Analysis:} To gain deeper insights into model behavior, we generated classification reports and confusion matrices for each fold. The confusion matrices were visualized to enable identification of systematic misclassifications. These matrices were averaged across folds to provide an aggregate view of model confusions.
\end{enumerate}
\subsubsection{Model Selection and Training Regimen}

We conducted experiments on both English and Spanish datasets. The specific models used for each language are detailed in Tables~\ref{tab:english_models} and~\ref{tab:spanish_models}. These models were selected based on their suitability and previous success in addressing similar NLP challenges as addressed in Section~\ref{Section: LR Method}. All models were fine-tuned using the AdamW optimizer with a learning rate of 3e-5 and a weight decay of 0.01. We trained for a maximum of 15 epochs per fold, employing early stopping with a patience of 3 epochs based on the performance on a held-out development set. The maximum sequence length was set to 100 tokens. A batch size of 32 was used for all experiments. The choice of these hyperparameters was guided by a coarse grid search on a separate development set, aiming to optimize performance while preventing overfitting.


\begin{table}[h!]
    \centering
    \caption{Pre-trained Transformer Models Used for English.}
    \label{tab:english_models}
    \begin{tabular}{ll}
        \toprule
        \textbf{Architecture} & \textbf{Model Name} \\
        \midrule
        Roberta & roberta-base \\
        Albert & albert-base-v2 \\
        Electra & google/electra-base-generator \\
        DistilBERT & distilbert-base-uncased \\
        \bottomrule
    \end{tabular}
\end{table}

\begin{table}[h!]
    \centering
    \caption{Pre-trained Transformer Models Used for Spanish.}
    \label{tab:spanish_models}
    \begin{tabular}{ll}
        \toprule
        \textbf{Architecture} & \textbf{Model Name} \\
        \midrule
        Roberta & PlanTL-GOB-ES/roberta-base-bne \\
        Albert & dccuchile/albert-base-spanish \\
        Electra & mrm8488/electricidad-base-generator \\
        DistilBERT & mrm8488/distill-bert-base-spanish-wwm-cased-finetuned-spa-squad2-es \\
        \bottomrule
    \end{tabular}
\end{table}

Through this fine-tuning exercise, we aim to establish strong, reproducible baselines for the PolyHope V2. The results obtained will quantify the effectiveness of Transformer models for hope speech detection and, critically, sarcasm identification. This comparative analysis will provide a benchmark against which to evaluate the performance gains achieved by our proposed ZSL and FSL methodologies leveraging LLMs.

\subsection{Zero-Shot and Few-Shot Learning with Large Language Models}

We further investigated the potential of LLMs to perform hope speech and sarcasm detection in ZSL and FSL settings. This approach allows us to leverage the pre-existing knowledge and reasoning capabilities of LLMs without extensive task-specific training. We employed GPT-4 and Llama 3, two state-of-the-art LLMs, to explore their effectiveness in this context.

We selected GPT-4~\cite{achiam2023gpt, bubeck2023sparks} and Llama 3~\cite{grattafiori2024llama} for their superior performance in various natural language understanding and generation tasks. GPT-4, known for its advanced reasoning and generalization abilities, serves as a high-performing benchmark. Llama 3, a powerful open-source LLM, offers an alternative perspective and allows for greater control and customization.

\subsubsection{Prompt Engineering}

The performance of LLMs in ZSL and FSL settings is highly dependent on the design of effective prompts. We carefully crafted prompts to elicit the desired classification behavior from the LLMs.

\textbf{Zero-Shot Prompts:}
For ZSL, we used a simple prompt that instructed the LLM to classify the input text into one of the predefined categories (Hope, Not Hope for binary classification; Generalized Hope, Realistic Hope, Unrealistic Hope, Not Hope, Sarcasm for multiclass classification). The prompt was formatted as follows:

\begin{verbatim}
Classify the following text into one of the categories \{categories\}:

Text: \{text\}
Label:
\end{verbatim}

\textbf{Few-Shot Prompts:}
For FSL, we provided the LLMs with a set of labeled examples to guide their classification. We selected 5 balanced examples per class. The prompt was structured as follows:

\begin{verbatim}
Here are some labeled examples:

Text: \{example1_text\}
Label: \{example1_label\}
Text: \{example2_text\}
Label: \{example2_label\}
...

Now classify this text:
Text: \{text\}
Label:
\end{verbatim}

\subsubsection{Implementation Details}

\textbf{GPT-4:}
We used the OpenAI API to access GPT-4. The \texttt{classify\_text} function sends a request to the OpenAI API with the appropriate prompt and receives the LLM's classification. The temperature parameter was set to 0 to encourage deterministic and consistent responses.

\begin{verbatim}
response = openai.chat.completions.create(
    model="gpt-4",
    messages=[{"role": "system", "content": "You are an NLP model trained
    for text classification."},
              {"role": "user", "content": prompt}],
    temperature=0
)
\end{verbatim}

\textbf{Llama 3:}
We accessed Llama 3 via the Together AI API. The \texttt{format\_few\_shot\_prompt} function constructs the prompt, including a system message defining the LLM's role as a text classification expert and providing example classifications. The Together AI API client then sends the prompt to the Llama 3 model. We set the temperature to 0.3 and the maximum number of tokens to 50.

\begin{verbatim}
response = client.chat.completions.create(
    model="meta-llama/Llama-3-70b-chat-hf",
    messages=messages,
    temperature=0.3,
    max_tokens=50,
    stop=["\n"],
    stream=True
)
\end{verbatim}

To evaluate the performance of both GPT-4 and Llama 3, we used the same evaluation metrics as in the fine-tuning experiments. These metrics allow for a direct comparison between the LLM-based approaches and the fine-tuned Transformer baselines.

By evaluating GPT-4 and Llama 3 in ZSL and FSL settings, we aim to assess the potential of LLMs for hope speech and sarcasm detection without task-specific training. For the FSL setting, we have taken a sample of 10 per label in our prompt draft and predicted on the rest of the datasets for both languages. The results will provide insights into the strengths and limitations of these models and inform future research directions in this area. We expect that FSL will outperform ZSL, and that GPT-4 and Llama 3 will exhibit different strengths and weaknesses due to their architectural differences and training data. A comparison with the fine-tuned Transformer baselines will reveal the trade-offs between these two approaches and highlight the potential for combining them to achieve even better performance.

\section{Results}

To evaluate the performance of both fine-tuned and prompt-based models on the PolyHope V2 dataset, we conducted experiments across binary and multiclass hope speech detection tasks in English and Spanish. The following tables present detailed results from two modeling strategies: (i) supervised fine-tuning of Pre-trained Language Models (PLMs), and (ii) prompt-based classification using large generative models in zero-shot and few-shot settings. Tables\ref{tab:ResPLM_EN} and~\ref{tab:ResPLM_ES} report the outcomes of PLM fine-tuning, while Tables~\ref{tab:Resprompt_en} and~\ref{tab:Resprompt_es} summarize the performance of GPT-4 and Llama-3 under prompt-based configurations. Each table includes weighted and macro-averaged precision, recall, F1-score, and overall accuracy to provide a comprehensive comparison across approaches, tasks, and languages.

\begin{table*}[ht]
\centering
\caption{\label{tab:ResPLM_EN}Results for English language by fine tuning PLMs.}
\scalebox{0.9}{
\begin{tabular}{l|ccc|ccc|c}
 \multirow{2}{*}{\bf Model} & \multicolumn{3}{c}{\bf Avg. Weighted Scores} & \multicolumn{3}{c}{\bf Avg. Macro Scores} & \multirow{2}{*}{\bf Accuracy} \\
        &Precision&Recall&F1-score&Precision&Recall&F1-score& \\

\hline
\multicolumn{8}{c}{\bf Binary Hope Speech Detection}\\
\hline

RoBERTa &0.8656&0.8650&0.8651&0.8643&0.8653&0.8646&0.8650 \\
Albert &0.8330&0.8327&0.8326&0.8326&0.8314&0.8317&0.8327 \\
Electra &0.8530&0.8525&0.8526&0.8517&0.8526&0.8520&0.8525 \\
DistilBERT &0.8604&0.8599&0.8599&0.8591&0.8601&0.8594&0.8599 \\
\hline
\multicolumn{8}{c}{\bf Multiclass Hope Speech Detection}\\
\hline
RoBERTa &0.7929&0.7885&0.7900&0.7551&0.7645&0.7587&0.7885 \\
Albert &0.7728&0.7741&0.7723&0.7474&0.7251&0.7347&0.7741 \\
Electra &0.7705&0.7626&0.7657&0.7217&0.7361&0.7277&0.7626 \\
DistilBERT &0.7837&0.7813&0.7822&0.7480&0.7450&0.7461&0.7813 \\ \hline

\end{tabular}
}

\end{table*}

\begin{table*}[ht]
\centering
\caption{\label{tab:Resprompt_en}Results for English language using prompt-learning.}
\scalebox{0.8}{
\begin{tabular}{l|l|ccc|ccc|c}
 \multirow{2}{*}{\bf Model} &\multirow{2}{*}{\bf Approach} & \multicolumn{3}{c}{\bf Avg. Weighted Scores} & \multicolumn{3}{c}{\bf Avg. Macro Scores} & \multirow{2}{*}{\bf Accuracy} \\
        &&Precision&Recall&F1-score&Precision&Recall&F1-score& \\

\hline
\multicolumn{9}{c}{\bf Binary Hope Speech Detection}\\
\hline

Llama3 & FSL & 0.7626 & 0.7401 & 0.7379 & 0.7566 & 0.7481 & 0.7391 & 0.7401 \\
       & ZSL & 0.6743 & 0.6695 & 0.6678 & 0.6728 & 0.6592 & 0.6603 & 0.6784 \\ \hline

GPT-4  & FSL & 0.7932 & 0.7789 & 0.7795 & 0.7861 & 0.7724 & 0.7711 & 0.7823 \\
       & ZSL & 0.7717 & 0.7706 & 0.7692 & 0.7724 & 0.7662 & 0.7674 & 0.7706 \\ 
\hline
\multicolumn{9}{c}{\bf Multiclass Hope Speech Detection}\\
\hline
Llama3 & FSL & 0.5090 & 0.4374 & 0.4244 & 0.4677 & 0.3933 & 0.3494 & 0.4374 \\
       & ZSL & 0.3886 & 0.3741 & 0.3698 & 0.3517 & 0.3086 & 0.2979 & 0.3772 \\\hline

GPT-4  & FSL & 0.4872 & 0.4594 & 0.4485 & 0.4388 & 0.4096 & 0.3963 & 0.4481 \\
       & ZSL & 0.5445 & 0.4354 & 0.4282 & 0.4638 & 0.4497 & 0.3690 & 0.4354 \\ \hline

\end{tabular}
}
\end{table*}

\begin{table*}[ht]
\centering
\caption{\label{tab:ResPLM_ES}Results for Spanish language by fine tuning PLMs.}
\scalebox{0.9}{
\begin{tabular}{l|ccc|ccc|c}
 \multirow{2}{*}{\bf Model} & \multicolumn{3}{c}{\bf Avg. Weighted Scores} & \multicolumn{3}{c}{\bf Avg. Macro Scores} & \multirow{2}{*}{\bf Accuracy} \\
        &Precision&Recall&F1-score&Precision&Recall&F1-score& \\

\hline
\multicolumn{8}{c}{\bf Binary Hope Speech Detection}\\
\hline
Roberta &0.8410&0.8401&0.8401&0.8403&0.8400&0.8397&0.8401 \\
Albert &0.8426&0.8425&0.8425&0.8420&0.8422&0.8420&0.8425 \\
Electra &0.8332&0.8308&0.8309&0.8314&0.8323&0.8308&0.8308 \\
DistilBERT &0.7699&0.7697&0.7697&0.7691&0.7692&0.7690&0.7697 \\

\hline
\multicolumn{8}{c}{\bf Multiclass Hope Speech Detection}\\
\hline
Roberta &0.7603&0.7604&0.7593&0.7343&0.7242&0.7281&0.7604 \\
Albert &0.7587&0.7585&0.7581&0.7306&0.7229&0.7261&0.7585 \\
Electra &0.7493&0.7410&0.7444&0.7038&0.7176&0.7097&0.7410 \\
DistilBERT &0.6501&0.6530&0.6509&0.6154&0.5986&0.6061&0.6530 \\ \hline

\end{tabular}
}

\end{table*}

\begin{table*}[ht]
\centering
\caption{\label{tab:Resprompt_es}Results for Spanish language using prompt-learning.}
\scalebox{0.8}{
\begin{tabular}{l|l|ccc|ccc|c}
 \multirow{2}{*}{\bf Model} &\multirow{2}{*}{\bf Approach} & \multicolumn{3}{c}{\bf Avg. Weighted Scores} & \multicolumn{3}{c}{\bf Avg. Macro Scores} & \multirow{2}{*}{\bf Accuracy} \\
        &&Precision&Recall&F1-score&Precision&Recall&F1-score& \\

\hline
\multicolumn{9}{c}{\bf Binary Hope Speech Detection}\\
\hline

Llama3 & FSL & 0.7415 & 0.7221 & 0.7194 & 0.7374 & 0.7283 & 0.7205 & 0.7221 \\
       & ZSL & 0.6627 & 0.6472 & 0.6431 & 0.6588 & 0.6410 & 0.6359 & 0.6489 \\ \hline

GPT-4  & FSL & 0.7756 & 0.7562 & 0.7547 & 0.7682 & 0.7497 & 0.7473 & 0.7554 \\
       & ZSL & 0.7284 & 0.7159 & 0.7123 & 0.7230 & 0.7102 & 0.7078 & 0.7167 \\ 
\hline
\multicolumn{9}{c}{\bf Multiclass Hope Speech Detection}\\
\hline
Llama3 & FSL & 0.5508 & 0.5225 & 0.5134 & 0.4647 & 0.3999 & 0.3756 & 0.5225 \\
       & ZSL & 0.3928 & 0.3645 & 0.3551 & 0.3457 & 0.3004 & 0.2819 & 0.3674 \\\hline

GPT-4  & FSL & 0.4993 & 0.4712 & 0.4639 & 0.4214 & 0.3937 & 0.3866 & 0.4712 \\
       & ZSL & 0.4547 & 0.4285 & 0.4196 & 0.4001 & 0.3792 & 0.3657 & 0.4285 \\ \hline

\end{tabular}
}
\end{table*}

Fine‑tuning domain‑specific transformers consistently delivers the strongest and most balanced performance across both languages (Tables \ref{tab:ResPLM_EN}, \ref{tab:ResPLM_ES}). In the binary task, English \textbf{RoBERTa} reaches a macro $F_{1}$ of 86.5\%, while its Spanish counterpart \textbf{Albert} attains 84.2\%; the weighted–macro divergences are $\leq0.3$\% in either language, confirming uniform treatment of \emph{Hope} and \emph{Not‑Hope}.  Scaling to five classes reduces every metric by roughly 10\% for English (RoBERTa macro $F_{1}$=75.9\%) and about 12\% for Spanish (Albert macro $F_{1}$=72.6\%), yet the weighted–macro gaps stay below 3\%, indicating that the extra difficulty is spread fairly rather than concentrating on the minority labels \emph{Sarcasm}, \emph{Unrealistic Hope}, and \emph{Realistic Hope}.  Prompt‑based large‑language‑model baselines (Tables \ref{tab:Resprompt_en}, \ref{tab:Resprompt_es}) narrow the margin in the binary setting—\textbf{GPT‑4} with ten in‑context demonstrations yields weighted $F_{1}$ scores of 77.9\% (English) and 75.5\% (Spanish) but still lag the best fine‑tuned transformer by 8–9\% and exhibit macro scores 5–9\% lower than their own weighted figures, signalling over‑reliance on the dominant \emph{Not‑Hope} class.  Zero‑shot prompting widens these disparities further, bringing five‑fold macro $F_{1}$ down to 36–40\% in both languages.  Overall, the empirical evidence shows that supervised transformer adaptation not only yields higher aggregate accuracy ($\approx$79\% English, $\approx$73\% Spanish, five‑fold) but also preserves class balance, whereas few‑ and zero‑shot prompting—while informative as a reasoning baseline—remains insufficient for the fine‑grained distinctions mandated by the PolyHope~V2 dataset.

Several interacting factors explain the systematic gap observed in Tables \ref{tab:Resprompt_en}–\ref{tab:Resprompt_es}. First, the transformers were \emph{discriminative models} explicitly fine‑tuned on PolyHope tweets; gradient updates aligned their internal representations with the lexical idiosyncrasies of X/Twitter, the pragmatic cues that mark sarcasm, and the subtle distinctions among our three hope sub‑types. GPT‑4 and Llama-3, by contrast, were deployed in a zero‑ or few‑shot generative regime: without parameter adaptation they relied on priors learned from broad, largely formal corpora where the boundary between ``realistic’’ and ``unrealistic’’ hope is seldom annotated. Second, the demonstration pool was deliberately kept minimal (ten examples per class) to probe “out‑of‑the‑box” reasoning. This is dwarfed by the stylistic diversity and label noise of social‑media text, so the LLMs struggle to infer robust class‑conditional cues, particularly for the minority labels that constitute $<10$\% of either language’s corpus. 

Fine‑tuned Pre-trained Language Models (PLMs), in contrast, see every training instance and therefore learn token‑level features even for sparse categories. Third, prompt‑based classification is highly sensitive to surface form: small edits in wording or label order can shift accuracy by more than 5\% \citep{zhou2022large}. Our study used a single handcrafted template to ensure comparability, but that template may sit at a sub‑optimal point in the LLM’s likelihood landscape. Supervised Pre-trained Language Models (PLMs) avoid this volatility by emitting logits directly over a fixed label vocabulary.
Moreover, the generative output must be parsed back into a discrete class, introducing an additional failure channel for example, returning \textit{wishful thinking} instead of \textit{Unrealistic Hope} or attaching extraneous punctuation which the evaluation counts as error even when the underlying reasoning is partially correct.

Finally, sarcastic hope often relies on culture specific pragmatic signals that differ between English and Spanish; without fine‑tuning, the LLMs lack exposure to these bilingual nuances, reducing macro scores in both languages. Hence, the observed deficiencies do not necessarily imply limited reasoning capacity in GPT‑4 or Llama 3; rather, they reflect domain mismatch, extremely small in‑context support, prompt sensitivity, and generative‑to‑discriminative conversion overheads—constraints we accepted so that the LLM runs would serve as a clean baseline of \emph{unadapted} reasoning on this complex multilingual taxonomy.

\section{Discussion and Analysis }

\begin{figure}
    \centering
    \includegraphics[width=0.7\linewidth]{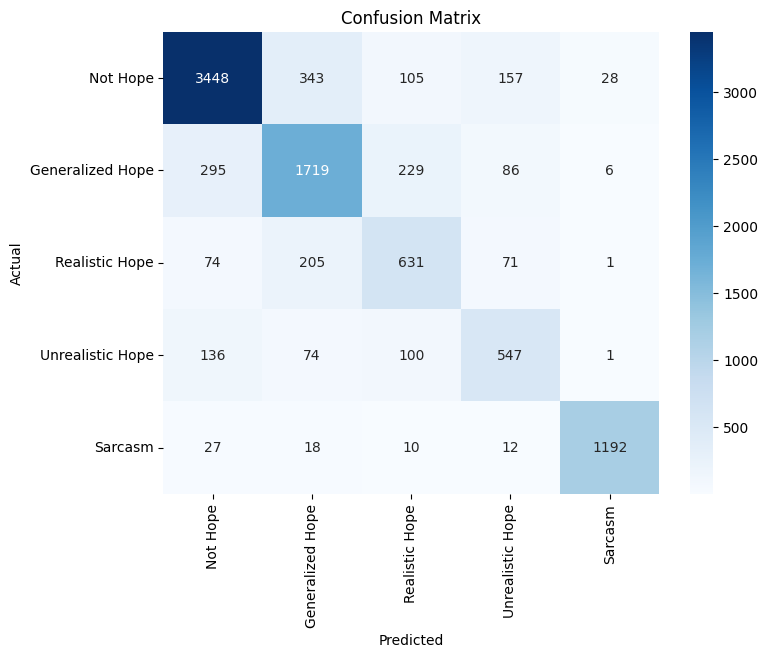}
    \caption{The figure shows the confusion matrix of the best-performing transformer model (RoBERTa) in English.}
    \label{fig:confusion_matrix_English}
\end{figure}

\begin{figure}
    \centering
    \includegraphics[width=0.7\linewidth]{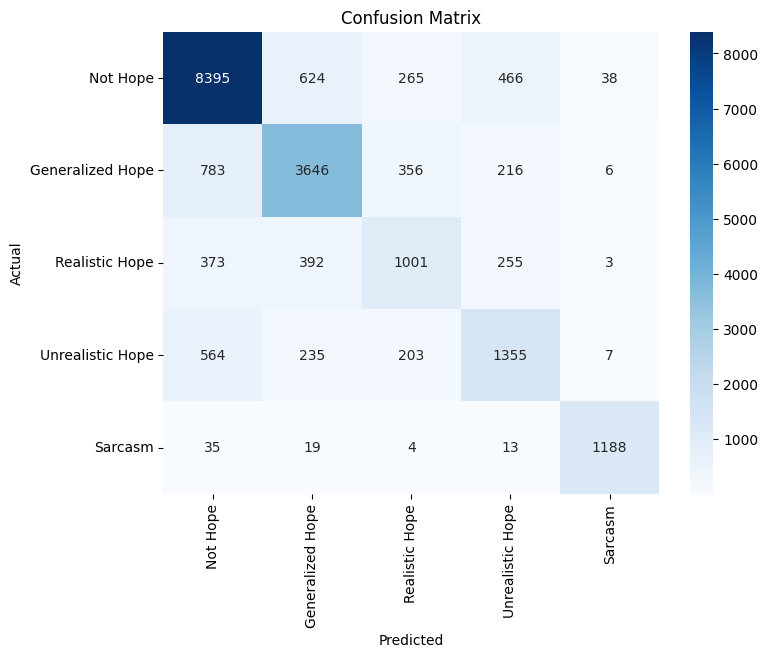}
    \caption{The figure shows the confusion matrix of the best-performing transformer model (RoBERTa) in Spanish.}
    \label{fig:confusion_matrix_Spanish}
\end{figure}

Figure~\ref{fig:confusion_matrix_English} and~\ref{fig:confusion_matrix_Spanish} shows the multi-class confusion matrix between the best performing transfomer models. Figure~\ref{fig:confusion_matrix_English} (English) and Figure~\ref{fig:confusion_matrix_Spanish} (Spanish) show that the best transformer already separates the two polar classes extremely well—\emph{Sarcasm} is identified in more than 94\% of cases in both languages, and \emph{Not‑Hope} achieves recalls of 85.4\% (3,448/4,081) in English and 85.8\% (8,395/9,788) in Spanish—yet the model falters when deciding \emph{which} variety of sincere hope is being expressed. In English almost one quarter of all \emph{Generalized Hope} tweets (295/2,335, 12.6\%) are downgraded to \emph{Not‑Hope}, while the three hope sub‑types frequently trade places with one another (e.g., 20.9\% of \emph{Realistic Hope} is predicted as \emph{Generalized}, 11.7\% of \emph{Unrealistic Hope} as \emph{Realistic}). The Spanish matrix magnifies the same phenomenon: although \emph{Generalized Hope} still attains a respectable 72.8\% recall (3,646/5,007), the model misroutes nearly one fifth of those instances to \emph{Not‑Hope} and, most strikingly, recognises fewer than half of the \emph{Realistic Hope} tweets correctly (49.5\%, 1,001/2,024), dispersing the remainder almost evenly across the other hope labels. These error patterns suggest that the network has learned robust cues for broad polarity and for pragmatic irony, but lacks the contextual or world‑knowledge signals required to gauge how grounded a hopeful statement is, a limitation that grows sharper under the stronger class imbalance of the Spanish data.

To understand \emph{why} the model stumbles on the three fine‑grained hope labels, we manually inspected 385 mis‑classified tweets from one English fold ($\approx 20\% $ of its 1,903 test items). Three recurrent patterns emerged:

\begin{itemize} \item \textbf{Lexical ‘‘hope’’ triggers without genuine optimism.} Many \emph{Not‑Hope} posts contain aspirational verbs in a negative frame, e.g.\ \emph{We \textbf{yearn} for change, but nothing ever improves''} (true Not‑Hope, predicted Generalized Hope). The vocabulary cue yearn’’ activates the hope prototype even though the surrounding discourse is pessimistic.

\item \textbf{Implicit stance overrides explicit markers.} Political commentary that affirms a group yet attacks its opponents is often annotated as \emph{Generalized Hope} but judged by the model as \emph{Not‑Hope}. Example: \emph{``\#Woke people accuse conservatives of narrow‑mindedness; I still \textbf{believe} we can find common ground’’} (true Generalized Hope, predicted Not-Hope). The optimism is subtle, buried inside an adversarial sentence.

\item \textbf{Boundary fuzziness between hope sub‑types.} Tweets that express a concrete but weakly grounded desire swing between \emph{Generalized}, \emph{Realistic}, and \emph{Unrealistic Hope}. For instance, \emph{Hey \#USER, \textbf{hope} you’ll read this on air tomorrow!''} (true Generalized Hope, predicted Realistic Hope) lacks external evidence and was annotated as \emph{Generalized}, yet the model judged the request sufficiently specific to be \emph{Realistic}. Conversely, progress‑update posts such as \emph{Still trying to finish Chapter 5 before July; \textbf{wish} me luck''} (true Realistic Hope, predicted Generalized Hope) are grounded in an attainable goal but resemble generic motivational tweets in form. \end{itemize}

These observations align with the quantitative error matrix: the classifier already recognises overt polarity and sarcastic cues, but its token‑level heuristics cannot reliably infer the speaker’s \emph{degree of realism}. Future work should therefore incorporate temporal or evidential features—e.g.\ futurity markers, probability adverbs, or external knowledge about plausibility—to sharpen the distinctions among \emph{Generalized}, \emph{Realistic}, and \emph{Unrealistic Hope}.

\section{Conclusion}
This paper introduces PolyHope V2, a novel bilingual dataset and modeling framework aimed at detecting fine-grained categories of hope speech—including sarcasm—across English and Spanish tweets. Our results demonstrate that fine-tuned transformer models consistently outperform zero-shot and few-shot large language models, particularly in maintaining class balance and accurately distinguishing among nuanced subtypes of hope. While LLMs like GPT-4 and Llama 3 show promise in binary classification tasks, their performance declines sharply in more granular multi-class settings due to their lack of domain adaptation and sensitivity to prompt design. The inclusion of sarcasm as a separate category further sharpens the challenge, revealing the limitations of current models in interpreting pragmatic cues. Our analysis underscores the need for richer annotations, cultural grounding, and external knowledge integration to advance the state of hope-speech detection. PolyHope V2 thus provides a strong empirical and methodological basis for future multilingual emotion detection research, bridging the gap between surface-level sentiment and deeper, context-aware emotional understanding.










\bibliography{sample.bib}

\end{document}